%% file: main.tex
\documentclass[conference]{IEEEtran}
\usepackage{rss}
\usepackage{tabu}
\usepackage{booktabs}
\usepackage{siunitx}
\usepackage{caption}
\usepackage{bm}
\usepackage{mathtools,amsmath} %
\usepackage{amssymb}  %
\usepackage{subcaption}
\usepackage{amsmath,mathtools}
\usepackage{soul}
\usepackage[normalem]{ulem}
\usepackage{caption}
\definecolor{somegray}{rgb}{0.5, 0.5, 0.5}
\newcommand{\darkgrayed}[1]{\textcolor{somegray}{#1}}
\makeatletter
\newcommand*\titleheader[1]{\gdef\@titleheader{#1}}
\AtBeginDocument{%
  \let\st@red@title\@title
  \def\@title{%
    \vskip-2.3em
    \bgroup\normalfont\large\centering\@titleheader\par\egroup
    \vskip1em\st@red@title}
}
\makeatother
\titleheader{\darkgrayed{This paper has been accepted for publication at Robotics: Science and Systems 2024 conference.}}

\newcolumntype{C}[1]{>{\centering}m{#1}}

\newcommand{\hide}[1]{}
\newcommand{\lb}[1]{\underbar{$#1$}}
\newcommand{\ub}[1]{\overline{#1}}
\let\vec\bm
\newcommand{\mat}[1]{\begin{bmatrix}#1\end{bmatrix}}

\DeclareMathOperator*{\argmin}{argmin}

\newtheorem{proposition}{Proposition}

\newcommand{\rebuttal}[1]{{#1}}

\newcommand{\remove}[1]{}

\title{MPCC++: Model Predictive Contouring Control for Time-Optimal Flight with Safety Constraints}

\begin{document}

\author{\authorblockN{Maria Krinner,\authorrefmark{1}$^{1}$
Angel Romero,\authorrefmark{1}$^{2}$
Leonard Bauersfeld,$^{2}$ 
Melanie Zeilinger,$^{1}$
Andrea Carron,$^{1}$
Davide Scaramuzza$^{2}$
}
\authorblockA{
$^{1}$ Institute for Dynamics Systems and Control, ETH Zurich, Switzerland\\
$^{2}$ Robotics and Perception Group, University of Zurich, Switzerland\\
\authorrefmark{1}These authors contributed equally\vspace{15pt}
}
}

\makeatletter
\g@addto@macro\@maketitle{
  \captionsetup{type=figure}\setcounter{figure}{0}
  \centering
  \includegraphics[width=0.85\linewidth, height=7.0cm]{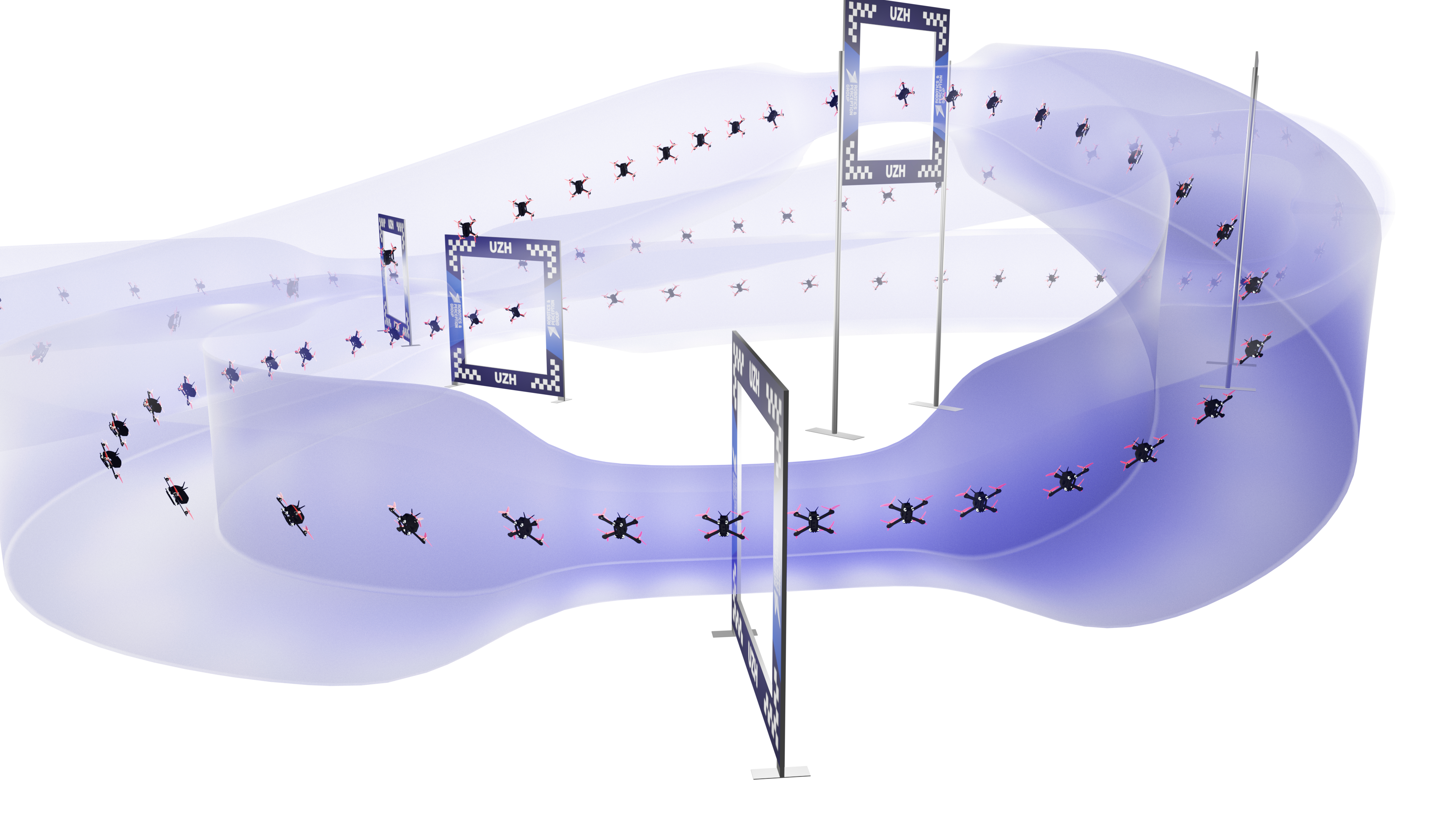}
  \captionof{figure}{The proposed method is able to fly a quadrotor at the limit of handling while ensuring safety. In the figure, the tunnel constraint is shown in translucent blue, and it expands in gate-free regions and shrinks at the gate passages, avoiding gate collisions. We are able to fly at speeds of more than 80 km/h while consistently preventing gate collisions and achieving 100\% success rate in real-world experiments.\vspace{-16pt}}
	\label{fig:catcheye}}
	\vspace{-4pt}
\makeatother

\maketitle

\input{sections/abstract}

\input{sections/introduction}
\input{sections/related_work}

\input{sections/methodology}
\input{sections/experiments}

\bibliographystyle{ieeetr}
\bibliography{references}

\end{document}

%% file: sections/abstract.tex
\begin{abstract}
Quadrotor flight is an extremely challenging problem due to the limited control authority encountered at the limit of handling.
Model Predictive Contouring Control (MPCC) has emerged as a \remove{leading}\rebuttal{promising} model-based approach for time optimization problems such as drone racing. 
However, the standard MPCC formulation used in quadrotor racing introduces the notion of the gates directly in the cost function, creating a multi-objective optimization that continuously trades off between maximizing progress and tracking the path accurately.
This paper introduces three key components that enhance the \rebuttal{state-of-the-art} MPCC approach for drone racing.
First and foremost, we provide safety guarantees in the form of a \rebuttal{track} constraint and terminal set.
The \remove{safety set}\rebuttal{track constraint} is designed as a spatial constraint which prevents gate collisions while allowing for time\remove{-} optimization only in the cost function.
Second, we augment the existing first principles dynamics with a residual term that captures complex aerodynamic effects and thrust forces learned directly from real-world data.
Third, we use Trust Region Bayesian Optimization (TuRBO), a state-of-the-art global Bayesian Optimization algorithm, to tune the hyperparameters of the \remove{MPC}\rebuttal{MPCC} controller given a sparse reward based on lap time minimization.
The proposed approach achieves similar lap times to the best\remove{ state-of-the-art}\rebuttal{-performing} RL \rebuttal{policy} and outperforms the best \remove{time-optimal}\rebuttal{model-based} controller while satisfying constraints.
In both simulation and real world, our approach consistently prevents gate crashes with 100\% success rate, while pushing the quadrotor to its physical limits reaching speeds of more than 80km/h. 

\end{abstract}

\section*{Supplementary Material}
\noindent A narrated video with real-world experiments is available at: \\
\url{https://youtu.be/sbKe9emghtM}

%% file: sections/introduction.tex
\section{Introduction}
The last decade has seen a remarkable growth in the number of quadrotor applications.
Many missions are time-critical, such as search and rescue, aerial delivery, flying cars, or space exploration~\cite{Loianno2020jfr, air_taxi_2020, shakhatreh2019unmanned}.
Among them, drone racing has emerged as a testbed for research on time-optimal flight.
Quadrotors are carefully engineered to push the limits of speed and agility, making drone racing an ideal benchmark for aerial performance and \remove{robustness}\rebuttal{safety} at high speeds~\cite{betz2022autonomous, hanover2023autonomous}.
State-of-the-art approaches to autonomous drone racing can be categorized into learning-based and optimization-based methods.

Within learning-based methods, Reinforcement Learning (RL) has arisen as an attractive alternative to conventional planning and control algorithms, outperforming world-champion pilots %
\cite{kaufmann23champion, Song23Reaching}.
\remove{Unlike optimal control, RL uses data sampled through millions of trial-and-error interactions with the environment to optimize a controller.}
\rebuttal{Unlike optimal control, RL optimizes a controller using data sampled from numerous trial-and-error interactions with the environment}. \remove{RL}\rebuttal{This approach} can manage sparse objectives and unstructured observation spaces, providing substantial flexibility and versatility in the controller design. %
As shown in \cite{Song23Reaching}, RL can directly optimize a task-level objective, eliminating the need for explicit intermediate representations such as trajectories. 
Furthermore, \remove{RL can leverage} \rebuttal{it leverages} domain randomization in simulation to cope with model uncertainty, allowing the discovery of more robust control responses. 
These advantages have facilitated numerous breakthroughs in pushing quadrotor systems to their operational limits\remove{ through RL \mbox{\cite{kaufmann23champion, Song23Reaching}}}.
However, despite \remove{the}\rebuttal{its} empirical success, RL \remove{controllers}\rebuttal{policies} lack theoretical guarantees.
Integrating safety considerations into \remove{machine learning}\rebuttal{learning-based} frameworks\remove{, especially within RL,} remains an area of ongoing research \cite{gu2022reviewsafeRL, brunke2022safe}.
The combination of optimization-based and learning-based architectures to enhance safety \remove{in RL} is emerging as a prominent topic within the robotics community \cite{wabersich2021predictive, Lyapunovstable, romero2023actor}.

\remove{Conversely, optimization} \rebuttal{Optimization}-based architectures have tackled the problem from a different perspective. 
In~\cite{CPC}, the authors propose a method to find time-optimal trajectories through a predefined set of waypoints and show how they outperform expert drone-racing pilots. 
However, these trajectories take several hours to calculate, rendering them impractical for replanning in the face of model mismatch and unknown disturbances (e.g., drone model, gate positions, aerodynamic effects, wind gusts).
Addressing this challenge necessitates an algorithm capable of generating time-optimal trajectories in real-time.
In~\cite{mpcc, replanning}, a Model Predictive Contouring Control (MPCC) method is introduced to track non-feasible paths \remove{in a time-optimal fashion.} \rebuttal{by maximizing progress along the designated path in a manner akin to time-optimal strategies.}
By solving the complex time-allocation problem online at every time step, MPCC selects the optimal states and inputs that \remove{maximize the progress along the designated path. This approach yields a controller that shows great adaptation against model mismatch and unknown disturbances.} \rebuttal{enhance progress. This approach results in a controller that demonstrates great adaptatability in response to model mismatches and unknown disturbances.}

Several modifications to the MPCC cost function are necessary to adapt the \remove{MPCC} formulation to the drone racing task.
\remove{
First, the notion of gates is introduced in the objective function by parameterizing the contour weight as a function of the progress. This comes with adding extra hyperparameters, which i) introduces a multi-objective cost function that forces a sub-optimal trade-off between performance and safety, ii) does not generalize to different gate configurations, and iii) requires complex tuning due to the increased number of parameters.
Various strategies have been devised to address the tuning problem for contouring controllers.
For instance, \mbox{\cite{wml}} employs a policy search technique to explore the high-level parameter space of the cost function.
Similarly, \mbox{\cite{frohlich2021model}} explores the parameter space using local Bayesian Optimization (BO).
}
\rebuttal{
In~\cite{mpcc}, the concept of gates is introduced in the cost function by parameterising the contour weight as a multivariate Gaussian function. Specifically, this adaptation employs a Gaussian function at each gate to scale the contour weight in the proximity of the gates. This ensures that the drone closely follows the reference path when navigating through the gate. However, this approach presents several shortcomings. First, employing a multivariate Gaussian to model the contour weight significantly increases the number of hyperparameters, making manual tuning a highly challenging task \cite{wml}. 
Secondly, while this contour function encourages the drone's proximity to the designated path, it does not explicitly prevent gate collisions, resulting in a sub-optimal trade-off between performance and safety. Moreover, the contour function requires specific tuning for each gate configuration, which restricts its flexibility when gate positions change and increases the dimensionality of the parameter space with the number of gates.
}

Although this method has resulted in control capabilities surpassing those of human pilots, the approximations in the cost function are\remove{ still} exceedingly intricate, suggesting the possibility for more robust and scalable alternatives. Such alternatives would ideally achieve the desired outcomes by imposing constraints that consistently prevent gate collisions, avoiding the need to handcraft complex, high-parametric cost functions.

Previous works have aimed to tackle this issue by introducing a \remove{tunnel-shaped constraint set}\rebuttal{spatial constraint} around the trajectory of the quadrotor \mbox{\cite{majumdar2017funnel, ji2021cmpcc, arrizabalaga2022towards}}.
In particular, \mbox{\cite{arrizabalaga2022towards}} \remove{presents a tunnel-shaped constraint set by adapting}\rebuttal{adapts} the quadrotor's dynamics to a local Frenet-Serret\remove{ reference} frame, which naturally gives rise to \rebuttal{tunnel-like} position constraints\remove{ in the shape of a tunnel}.
\rebuttal{However, this spatial reformulation of the dynamics requires using first and second derivatives of the coordinates in the Frenet-Serret frame, which introduces singularities that require careful handling.}
\remove{The authors validate their approach in simulation and show that they can fly different tracks at high speeds within the confines of the tunnel.} %

\subsection*{Contributions}
This paper introduces three key components that enhance the state-of-the-art MPCC formulation for drone racing~\cite{mpcc}.
First, we provide safety guarantees in the form of a \rebuttal{track} constraint and terminal set.
The \remove{safety set}\rebuttal{track constraint} is designed as a spatial constraint that prevents gate collisions in the form of a prismatic tunnel\remove{, similar to \mbox{\cite{arrizabalaga2022towards}}}.
\rebuttal{Unlike \cite{arrizabalaga2022towards}, our formulation retains the dynamics in their standard Euclidean form, avoiding the complex reformulation into the Frenet-Serret frame and its associated singularities.}
The terminal set consists of a periodic, feasible \remove{centerline}\rebuttal{trajectory}.
This combination provides guarantees of recursive feasibility and inherent robustness.
Second, we augment the existing first principles dynamics with a residual term that captures complex aerodynamic effects and thrust forces \remove{learned}\rebuttal{inferred} directly from real-world data.
\rebuttal{Third, we show that using Trust-Region Bayesian Optimization (TuRBO)\remove{, a state-of-the-art Bayesian Optimization (BO) algorithm,} to tune the hyperparameters of the MPCC controller results in superior performance compared to previous work \cite{wml}.}
\remove{Third, we use Trust-Region Bayesian Optimization (TuRBO), a state-of-the-art global Bayesian Optimization (BO) algorithm, to tune the hyperparameters of the \remove{MPC}\rebuttal{MPCC} controller given a sparse reward based on lap time minimization.}
In both simulation and real-world experiments, we illustrate how combining these elements improves the controller's performance and \remove{robustness}\rebuttal{safety} beyond the state-of-the-art MPCC.
Moreover, the performance of our method aligns with that of the best-performing RL \remove{controller}\rebuttal{policy}, with the added benefit of incorporating safety into our formulation.

%% file: sections/related_work.tex
\section{Related Work}
\subsection{High-speed quadrotor flight}
The literature on high-speed\remove{, highly agile} quadrotor flight is categorized into two primary approaches: model-based and learning-based. A comprehensive survey of the literature on this subject can be found in \cite{hanover2023autonomous}.

The model-based category originates from polynomial planning and classical control techniques.
Traditionally, these methods focused on harnessing the differential flatness property of quadrotors and leverage the use of polynomial for planning ~\cite{Mueller11iros, Mahony12ram, Mellinger12ijrr, Mueller13iros}.
More recently, optimization-based methods have achieved planning of time-optimal trajectories using the quadrotor dynamics and numerical optimization~\cite{CPC}. 
Nonetheless, the substantial computational demand of these methods typically necessitates offline trajectory computation, rendering these approaches impractical for real-time applications.
\remove{The leading model-based solution for minimum-time flight is Model Predictive Contouring Control~(MPCC), which synthesizes the trajectory planning and control tasks into one.}
\rebuttal{Within model-based approaches, the problem is typically framed either as time minimization or as progress maximization, depending on the optimization problem's cost function. Time minimization formulations incorporate the time variable directly into the cost function but often require a spatial reformulation using the Frenet-Serret frame. This introduces several complexities, such as handling the inherent nonlinearities and singularities that arise from the coordinate transformations~\cite{spedicato, verscheure, pascoal}. Conversely, contouring methods propose employing progress maximization along a path as a proxy for time minimization. This approach simplifies the control problem by allowing for more robust and efficient solutions. This distinction underlines that although progress maximization does not address the time-optimal problem directly, it offers a practical alternative with significant success. Consequently, MPCC has shown promising results in achieving minimum-time flight~\cite{mpcc}.}

A significant benefit of optimization-based methods is their ability to incorporate safety \rebuttal{considerations} through state and input constraints.
Notably, works like ~\cite{majumdar2017funnel, ji2021cmpcc, arrizabalaga2022towards, qin2023time} have investigated the application of positional constraints in the form of gates or tunnels to prevent collisions.

On the other hand, a collection of\remove{ agile flight} learning-based methodologies for autonomous racing have emerged, which aim to replace the traditional planning and control layers with a neural network~\cite{Loquercio19TRO,kaufmann2022benchmark, rojas2020deeppilot}.
These purely data-driven control strategies, such as model-free \remove{reinforcement learning}\rebuttal{RL}, strive to circumvent the limitations of model-based controllers by learning effective \remove{controllers}\rebuttal{policies} directly from interactions with the environment.
For instance, \remove{Hwangbo et al.}\rebuttal{the authors in}~\cite{hwangbo2017control} \remove{demonstrated the application of}\rebuttal{employ} a neural network policy for guiding a quadrotor through waypoints and recovering from challenging initialization setups.  
\remove{Koch et al.}\rebuttal{In}~\cite{koch2019reinforcement}\remove{employed}, model-free RL \rebuttal{is employed} for low-level attitude control, \remove{and showed}\rebuttal{showing} that a learned low-level controller trained with Proximal Policy Optimization (PPO) outperformed a fully tuned PID controller\remove{ on almost every metric}.
\remove{Lambert et al.}\rebuttal{In}~\cite{lambert2019low}\remove{ implemented}, model-based RL \rebuttal{is used} to train a hovering controller. 
The family of learning-based approaches is rapidly advancing, fueled by recent breakthroughs in quadrotor simulation environments. These advancements have resulted in test environments that facilitate the training, assessment, and zero-shot transfer of control policies to the real world.
The state-of-the-art on realistic simulations for quadrotors is \cite{bauersfeld2021neurobem}, which introduces a hybrid aerodynamic quadrotor model that combines \remove{blade element momentum theory}\rebuttal{Blade Element Momentum Theory (BEM)} with learned aerodynamic representations from highly aggressive maneuvers.

\remove{Song et al.}\rebuttal{The authors of}~\cite{song2021autonomous} employed deep RL to generate near time-optimal trajectories for autonomous drone racing and tracked the trajectories by an MPC controller. 
More recently, \cite{kaufmann23champion, Song23Reaching} have demonstrated that policies trained with model-free RL can achieve super-human performance\remove{ at drone racing}. 
However, none of these learning\remove{-driven}\rebuttal{-based} approaches include safety considerations into their designs.
\rebuttal{
\subsection{Tunnel MPC}
The concept of employing spatial constraints to systematically prevent collisions is well-established in car racing, where tracks are clearly defined at all points. However, drone racing introduces a unique challenge: while the positions of the gates are fixed, the space in between remains open for exploration. Recent advancements in this field have introduced the idea of using a tunnel around the trajectory as a spatial constraint \cite{majumdar2017funnel, ji2021cmpcc, arrizabalaga2022towards}. This method involves a spatial reformulation of the drone's dynamics into the Frenet-Serret frame, focusing on the transverse distances from the drone's current position to the track's centerline. Such a state definition, while providing a natural mechanism to enforce tunnel constraints, necessitates a conversion of Euclidean coordinates into terms dependent on these transverse distances. Although this approach elegantly integrates the spatial constraints within the drone's control framework and has shown promising results in simulations~\cite{arrizabalaga2022towards}, including high-speed navigation across various tracks, it also presents significant challenges due to the spatial reformulation required. Specifically, it introduces singularities that must be carefully addressed to maintain a smooth and reliable flight trajectory.
}
\subsection{Data-driven models for MPC}
Learning-based MPC~\cite{spielberg2021neural, chee2022knode, saviolo2022physics, williams2018information, ostafew2016robust, rosolia2019learning, torrente2021data, mehndiratta2018automated, carron2019data} utilize real-world data to refine dynamic models or learn an objective function tailored for MPC applications. 
These strategies typically focus on learning dynamics for tasks where deriving an analytical model of the robot or their operational environments poses significant challenges. This is particularly relevant for highly dynamic tasks, such as aggressive autonomous driving around a loose-surface track, \remove{expemplified}\rebuttal{exemplified} in~\cite{williams2017information}.

Another promising line of research\remove{\mbox{~\cite{lahr2023zero}}} is the development of specialized solvers designed for use within a learning-based MPC framework\rebuttal{~\cite{lahr2023zero}}. This direction takes advantage of zero-order Sequential Quadratic Programming (SQP) techniques, potentially enhancing the efficiency\remove{ and effectiveness} of the control solutions for complex\remove{, dynamic} tasks. This includes adapting to unpredictable elements and optimizing performance over a wide range of operational conditions.
\subsection{Automatic controller tuning}
The classic approach \cite{MITrule} for controller tuning analytically finds the relationship between a performance metric, e.g. tracking error or trajectory completion, \rebuttal{and the controller parameters.} \remove{and}\rebuttal{It then} optimizes the parameters with gradient-based \remove{optimization}\rebuttal{methods} \cite{grimble1984implicit, aastrom1993automatic, mohd2015intelligent}.
However, expressing the long term \remove{measure}\rebuttal{performance metric}\remove{(in our case the laptime and gate passing metrics)}\rebuttal{, such as the lap time,} as a function of the tuning parameters is impractical and generally intractable.
There exist different methods for automatically tuning controllers \cite{schperberg2022auto, zanon2020safe, cheng2022difftune, tunempc, loquercio2022autotune, edwards2021automatic}.
\remove{
Another line of work proposes to iteratively estimate the optimization function, and use the estimate to find optimal parameters \mbox{\cite{menner2020maximum, berkenkamp2016safe, marco2016automatic}}.
There have also been RL-rooted methods to find the right set of hyperparameters for a controller \mbox{\cite{wml, song2021autonomous}}.
Other methods are rooted in Bayesian Optimization, such as \mbox{\cite{frohlich2021model, turbo}}.
}
\rebuttal{
RL methods~\cite{song2021autonomous} utilize trial-and-error to adapt controller settings to complex system dynamics without the need for explicit modeling.
Similarly, BO~\cite{frohlich2021model} leverages probabilistic models to navigate and explore the parameter space of black-box functions, effectively identifying optimal settings with minimal data.
Furthermore, Weighted Maximum Likelihood Estimation (WML)~\cite{wml} employs policy search techniques to iteratively refine parameter estimates from observed data, aiming to pinpoint the most effective parameters for the cost function.
These methods provide a range of techniques for controller tuning, each with its own set of strengths and limitations.
}

%% file: sections/methodology.tex
\section{Preliminaries}
\label{sec:prelim}
In this section we introduce the nominal quadrotor dynamics model \remove{that is later extended with learned dynamics. 
Additionally, we }\rebuttal{and} present the basics of the MPCC algorithm from \cite{mpcc}.
\subsection{Quadrotor Dynamics}
\label{sec:quad_dynamics}
In this section, we describe the nominal dynamics $\vec{f}(\vec{x}, \vec{u})$\remove{ used throughout this paper,} where $\vec{x} \in \mathbb{R}^{13}$ is the state of the quadrotor and $\vec{u} \in \mathbb{R}^4$ is the input to the system.
The state of the quadrotor \remove{described from the inertial frame $I$ to the body frame $B$.
Therefore,}\rebuttal{is given by} $\vec{x} = [\vec{p}_{I\remove{B}}, \vec{q}_{IB}, \vec{v}_{I\remove{I}}, \vec{w}_{B}]^T$, where $\vec{p}_{I\remove{B}}\in \mathbb{R}^3$ is the position, $\vec{q}_{IB} \in \mathbb{SO}(3)$ is the unit quaternion that describes the rotation \remove{of the platform}\rebuttal{from the body to the inertial frame}, $\vec{v}_{I\remove{I}} \in \mathbb{R}^3$ is the linear velocity vector, and $\vec{\omega}_{B} \in \mathbb{R}^3$ are the bodyrates in the body frame \rebuttal{$B$}.
\remove{
The input to the system is given as the collective thrust $\vec{f}_T = [0~~0 ~~f_{Bz}]^T$ and body torques $\vec{\tau}_T$.
}
For ease of readability, we drop the frame indices, as they \remove{are}\rebuttal{remain} consistent throughout the description.
The nominal dynamic equations are \rebuttal{given by:}
\begin{gather}
\begin{aligned}
\dot{\vec{p}} &= \vec{v} & \dot{\vec{v}} &= \vec{g} + \frac{\mathbf{R}(\vec{q}) \vec{f}_T}{m}\\
\dot{\vec{q}} &= \frac{\vec{q}}{2} \odot [0~~\vec{\omega}]^T &
\dot{\vec{\omega}} &= \mathbf{J}^{-1} \left( \vec{\tau}_T - \vec{\omega} \times \mathbf{J} \vec{\omega} \right)
\label{eq:quad_dynamics}
\end{aligned}
\end{gather}
where $\odot$ represents the Hamilton quaternion multiplication, $\mathbf{R}(\vec{q})$ the quaternion rotation, $m$ the quadrotor's mass, \remove{and} $\mathbf{J}$ the quadrotor's inertia\rebuttal{, $\vec{f}_T$ the collective thrust, and $\vec{\tau}_T$ the body torques}.
\remove{Additionally, the input space given by $\vec{f}$ and $\vec{\tau}$ is}\rebuttal{The input space, given by $\vec{f}_T$ and $\vec{\tau}_T$, is further} decomposed into single rotor thrusts $\vec{f} = [f_1, f_2, f_3, f_4]$ \remove{where $f_i$ is the thrust at rotor $i \in \{ 1, 2, 3, 4 \}$}\rebuttal{as follows:}
\begin{align}
\vec{f}_T &= \mat{0 \\ 0 \\ \sum f_i} &
\text{and }
\vec{\tau}_T &=
\mat{l/\sqrt{2} (f_1 + f_2 - f_3 - f_4) \\
l/\sqrt{2} (- f_1 + f_2 + f_3 - f_4) \\
c_\tau (f_1 - f_2 + f_3 - f_4)}
\label{eq:quad_inputs}
\end{align}
where \rebuttal{$f_i$ is the i-th rotor's thrust,} $l$ the quadrotor's arm length, and $c_\tau$ the rotor's torque constant.

In addition to \remove{these}\rebuttal{the} nominal dynamics, we introduce the full dynamic model $\vec{\hat{f}}(\vec{x}, \vec{u}) = \vec{f}(\vec{x}, \vec{u}) + \vec{g}(\vec{x}, \vec{u})$, which is augmented with a residual term $\vec{g}(\vec{x}, \vec{u})$ that captures unmodeled terms such as aerodynamic effects\remove{ and thrust forces}. The residual term is inferred \remove{using}\rebuttal{from} real-world data.

\subsection{Model Predictive Contouring Control}\label{sec:mpcc}
\remove{Consider}\rebuttal{We consider} the discrete-time dynamic system of a quadrotor with continuous state and input spaces, $\vec{x}_k \in \mathcal{X}$ and $\vec{u}_k \in \mathcal{U}$ respectively. 
\remove{Let us}\rebuttal{We} denote the time discretized evolution of the system \mbox{$\vec{\hat{f}} : \mathcal{X} \times \mathcal{U} \mapsto \mathcal{X}$} such that:
\begin{align}
\vec{x}_{k + 1} = \vec{\hat{f}}(\vec{x}_k, \vec{u}_k)
\end{align}
where the \remove{sub-}index $k$ \remove{is used to denote}\rebuttal{refers to the} states and inputs at time $t_k$.
The general Optimal Control Problem \rebuttal{(OCP)} considers the task of finding a control policy $\pi(\vec{x})$, a map from the current state to the optimal input, $\pi : \mathcal{X} \mapsto \mathcal{U}$, such that the cost function $J: \mathcal{X} \mapsto \mathbb{R}^+$ is minimized:
\begin{align}
\pi(\vec{x}) = & \argmin_u
  & & 
    J(\vec{x}) \notag\\
        & \text{subject to} && \vec{x}_0 = \vec{x} \notag \\ 
        &&& \vec{x}_{k+1} = \vec{\hat{f}}(\vec{x}_k, \vec{u}_k) \notag\\
        &&& \vec{x}_k \in \mathcal{X}, \quad \vec{u}_k \in \mathcal{U}
        \label{eq:ocp}
\end{align}
In MPCC, the goal is to compromise between maximizing progress along a predefined path, while tracking it accurately. The main ingredients of the cost function are the progress term $\theta$, the contour error $\vec{e^c}(\theta)$, and the lag error $\vec{e^l}(\theta)$, which describe the perpendicular and tangential error between the current position and its projection on the reference path:
\begin{align}
    \label{eq:J_mpcc}
    J_{MPCC}(\vec{x}) = \sum_{k = 0}^{N} \Vert \vec{e^l}(\theta_{k}) \Vert_{Q_l}^2 + \Vert \vec{e^c}(\theta_{k}) \Vert_{Q_c}^2 - \mu \rebuttal{v_{\theta_k}}
\end{align}
The resulting \remove{optimal control problem}\rebuttal{OCP} is formulated as follows:
\begin{align}
  \pi(\vec{x}) = & \argmin_u
  & & \begin{multlined}\sum_{k = 0}^{N } \Vert \vec{e^l}(\theta_k) \Vert_{Q_l}^2 + \Vert \vec{e^c}(\theta_k) \Vert_{Q_c}^2 + \Vert \vec{\omega_k}\Vert_{Q_{\omega}}^2 \\  + \Vert v_{\theta_k}\Vert_{R_{v_{\theta}}}^2 + \Vert\Delta \vec{f_k}\Vert_{R_{\Delta f}}^2 - \mu v_{\theta_k}
  \end{multlined} \notag\\
  & \text{subject to} & & \vec{x}_0 = \vec{x} \notag \\
  &&&\vec{x}_{k+1} = \vec{\hat{f}}(\vec{x}_k, \vec{u}_k) \notag \\
  &&&\lb{\vec{\omega}} \leq \vec{\omega} \leq \ub{\vec{\omega}} \notag \\
  &&&\lb{\vec{f}} \leq \vec{f} \leq \ub{\vec{f}} \notag \\
  &&& 0 \leq v_{\theta} \leq \ub{v_{\theta}} \notag \\
  &&& \lb{\Delta \vec{f}} \leq \Delta \vec{f} \leq \ub{\Delta \vec{f}}
  \label{eq:full_ocp}
\end{align}
where $v_{\theta}$ is the first derivative of $\theta$ with respect to time\remove{ , and $\vec{p}^d(\theta_N)$ is the terminal set which computation will be detailed in the next section}.
\remove{Note that we}\rebuttal{We} model the progress $\theta$ as a first order system\remove{ in order} to penalize the variation in progress in the cost, which provides a much smoother state. The norms of the form $\Vert \cdot \Vert_A^2$ represent the weighted Euclidean inner product $\Vert \vec{v} \Vert^2_A = \vec{v}^T A \vec{v}$. 
For more details about this formulation we refer the reader to \cite{mpcc}.

\section{Methodology}

In this section we introduce our enhancements \remove{of}\rebuttal{to} the standard MPCC formulation \remove{introduced}\rebuttal{described} in Section \ref{sec:mpcc}.
\subsection{Safety \rebuttal{constraints}}
\label{sec:tunnel_constraints}
\remove{In this section, we formally define}\rebuttal{We first introduce} the \remove{safety set}\rebuttal{track constraint} as a spatial constraint which consistently prevents gate collisions. We then define a\remove{ periodic} terminal set and show that the proposed controller is \remove{inherently robust}\rebuttal{recursively feasible}. 

\begin{figure}[tp]
\centering
\includegraphics[width=0.8\linewidth]{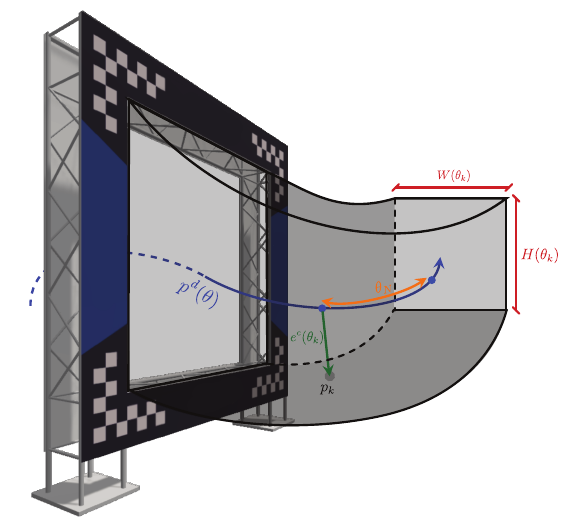}
\caption{\remove{Safety constraints in position in the shape of a tunnel. The size of the tunnel are set exactly the same as the dimensions of the interior of the gates where at the gates and larger everywhere else.}\rebuttal{Spatial constraint forming a tunnel around the centerline. The width and height are parameterized by $W(\theta_k)$ and $H(\theta_k)$, respectively.}}
\label{fig:tunnel}
\vspace{-15pt}
\end{figure}

We define the \remove{safety set}\rebuttal{track constraint} as a prismatic tunnel that joins the inner corners of the gates (see Figure \ref{fig:tunnel}). We refer to this formulation as MPCC++ to distinguish it from the baseline MPCC introduced in \cite{mpcc}. Two components are required to define such a set: i) a centerline which passes through the gate centers and matches the first derivative with the gate normal; \rebuttal{and} ii) a parameterization of the width and height of the cross section as a function of the progress $\theta$, i.e. $W(\theta_k)$, $H(\theta_k)$ respectively. 

We employ a simple hermetian spline as the centerline. The gate cross section is \remove{then patched}\rebuttal{constructed} around the centerline using standard Frenet-Serret formulas, similar to \cite{arrizabalaga2022towards}. At every point of the curve, consider the reference frame that consists of the vectors $[\vec{t}(\theta_k), \vec{n}(\theta_k), \vec{b}(\theta_k)]$.
Let $\vec{p}_k$ be the coordinates of the platform at current time $k$, and \remove{let} $\vec{p}^d(\theta_k)$\remove{ be} the \rebuttal{corresponding point on the} centerline. \remove{Given}\rebuttal{For} the width $W(\theta_k)$ and height $H(\theta_k)$ of the tunnel\remove{ at that point (as shown in Fig. \mbox{\ref{fig:tunnel})}, and given}\rebuttal{, define} the bottom left corner of the tunnel \rebuttal{as} $\vec{p}_0(\theta_k) = \vec{p}^d(\theta_k) - W(\theta_k) \cdot \vec{n}(\theta_k) - H(\theta_k) \cdot \vec{b}(\theta_k)$\remove{, we define the}\rebuttal{. The boundary of the tunnel is then determined by} four \remove{linear}\rebuttal{halfspace} constraints as \rebuttal{follows}:
\begin{align}
(\vec{p}_k - \vec{p}_0(\theta_k)) \cdot \vec{n}(\theta_k) \geq 0 \notag \\
2H(\theta_k) - (\vec{p}_k - \vec{p}_0(\theta_k)) \cdot \vec{n}(\theta_k) \geq 0 \notag \\
(\vec{p}_k - \vec{p}_0(\theta_k)) \cdot \vec{b}(\theta_k) \geq 0 \notag \\
2W(\theta_k) - (\vec{p}_k - \vec{p}_0(\theta_k)) \cdot \vec{b}(\theta_k) \geq 0
\label{eq:tunnel_constraints}
\end{align}
\remove{The constraints in \mbox{\eqref{eq:tunnel_constraints}} are added to problem \mbox{\eqref{eq:full_ocp}}, and are symbolically represented in Fig. \mbox{\ref{fig:tunnel}}.}
Furthermore, we set $W(\theta_k)=H(\theta_k)$ and parameterize \remove{the width and height of the tunnel}\rebuttal{$W(\theta_k)$} by \rebuttal{two distinct values:} a nominal value $W_n$ \rebuttal{for the broader sections of the tunnel,} and an inner gate value $W_{gate}$ \rebuttal{for the narrower sections at the gates}. A sigmoid function is employed to smoothly transition between these two levels\rebuttal{, ensuring a gradual narrowing of the tunnel as it approaches a gate}.

\remove{This terminal set is has been chosen to be a feasible trajectory that goes through the center of all gates:}
\remove{Such a set terminal set can be for example computed by solving the following optimization problem}
\rebuttal{We define the terminal set \rebuttal{$\mathcal{X}_f$} as a periodic feasible trajectory which passes through the center of the gates, similar to \cite{numerow2024inherently}:
\begin{align}
    \vec{p}_N \in \mathcal{X}_f
    \label{eq:terminal_constraint}
\end{align}
where $\vec{p}_N$ is the position of the last shooting node $\vec{x}_N$. By making the first and last elements of the trajectory coincide, we establish its periodicity, which ensures that the resulting trajectory is a positive invariant set. We compute this set offline by solving the following optimization problem:}
\begin{align}
& \argmin_{x,u}
  & & 
    \sum_{k=0}^M \|\vec{p}^d(\theta_k) - \vec{p}_k\|_2^2\notag\\
        & \text{subject to} && \vec{x}_0 = \rebuttal{\vec{x}_M} \notag \\ 
        &&& \vec{x}_{k+1} = \vec{\hat{f}}(\vec{x}_k, \vec{u}_k) \notag\\
        &&& \vec{x}_k \in \mathcal{X}, \quad \vec{u}_k \in \mathcal{U}
        \label{eq:ocp_proof}
\end{align}
\remove{where $M$ is the number of points that parameterize the terminal set, and the first constraints requires the trajectory to be periodic. The trajectory~$\vec{p}^d(\theta_k)$ is guaranteed to be positive invariant. Thanks to this terminal set, it is possible to state the following proposition.}
\rebuttal{where $M$ is the number of points that parameterize the terminal set, and the first constraint ensures periodicity of the trajectory. The resulting discrete terminal set is obtained from the solution of Problem \ref{eq:ocp_proof}:}
\begin{align}
    \mathcal{X}_f = \{ \vec{p}_0, \vec{p}_1, \cdots, \vec{p}_M\}
\end{align}
\rebuttal{where $\vec{p}_i$ is the position of the i-th shooting node.} 
\begin{proposition}
The MPCC formulation in \rebuttal{Problem}~\ref{eq:full_ocp} \rebuttal{subject to constraints~\eqref{eq:tunnel_constraints} and~\eqref{eq:terminal_constraint}} is recursively feasible and satisfies constraints at all times. 
\end{proposition}

The proof follows from standard MPC recursive feasibility arguments~\cite{rawlings2017model}. It is \rebuttal{also} possible to show that \remove{controller}\rebuttal{Problem}~\ref{eq:full_ocp} is\remove{ also} inherently robust to disturbances by following the same arguments used in~\mbox{\cite{numerow2024inherently}}.

\subsection{Dynamics augmentation}\label{sec:dynamics_augmentation}

In this section we \remove{explain how to} augment the nominal dynamics $\vec{f}(\vec{x}, \vec{u})$ introduced in Section \ref{sec:quad_dynamics}, with a residual term $\vec{g}(\vec{x}, \vec{u})$ that captures unmodeled effects.
We first provide a general overview of the forces and torques that act on the system, then identify the sources of model mismatch, and propose a set of polynomial features to approximate these terms from real-world data. 

We differentiate between the lift force $\vec{f}_{prop}$ produced by the propellers, and the collective aerodynamic forces $\vec{f}_{aero}$ which \remove{encompass}\rebuttal{encompasses} effects such as drag, induced lift, and blade flapping.
The total torque acting on the system is composed of four elements: \rebuttal{i)} the torque generated by the individual propellers $\vec{\tau}_{prop}$\remove{,}\rebuttal{; ii)} the yaw torque $\vec{\tau}_{mot}$ arising from changes in motor speeds\remove{,}\rebuttal{; iii)} the aerodynamic torque $\vec{\tau}_{aero}$ that captures a variety of aerodynamic influences\remove{, and}\rebuttal{; and iv)} \rebuttal{the} inertial term $\vec{\tau}_{iner}$. 

While $\vec{f}_{prop}$ and $\vec{\tau}_{prop}$ can be precisely estimated from first principles, \remove{accurately }modelling the aerodynamic \remove{forces}\rebuttal{terms} is significantly more challenging.
In \cite{bauersfeld2021neurobem}, \remove{the forces acting on the individual propellers}\rebuttal{$\vec{f}_{prop}$ and $\vec{\tau}_{prop}$} were modelled analytically using Blade Element Momentum Theory (BEM), which due to its precision, notably raises computational demands. 
\rebuttal{Instead, we follow the structure proposed by~\cite{kaufmann23champion}, in which a simple polynomial model is fitted to real-world data through conventional regression techniques. The polynomial model describes the aerodynamic forces and torques as a linear combination of polynomial features involving the drone's linear velocity (in body-frame) and the mean-squared rotor speed, $\Omega^2$.}
\vspace{-5pt}
\rebuttal{
\begin{align}
    f_{x} &= \vec{C}_{f_x} \begin{bmatrix} v_x & v_x^3 & \Omega^2 & v_x \Omega^2 \end{bmatrix}^T \notag \\
    f_y &= \vec{C}_{f_y} \begin{bmatrix} v_y & v_y^3 & \Omega^2 & v_y \Omega^2 \end{bmatrix}^T \notag \\
    f_z &= \vec{C}_{f_z} \begin{bmatrix} v_z & v_z^3 & v_{xy} & v_{xy}^2 & v_{xy} \Omega^2 & v_z \Omega^2 & v_{xy} v_z \Omega^2 \end{bmatrix}^T \notag \\
    \tau_x &= \vec{C}_{\tau_x} \begin{bmatrix} v_y & \Omega^2 & v_y \Omega^2 \end{bmatrix}^T \notag \\
    \tau_y &= \vec{C}_{\tau_y} \begin{bmatrix} v_x & \Omega^2 & v_x \Omega^2 \end{bmatrix}^T \notag \\
    \tau_z &= \vec{C}_{\tau_z} \begin{bmatrix} v_x & v_y \end{bmatrix}^T
\end{align}
}
\rebuttal{where $v_{xy}$ represents the horizontal velocity component. The selection of these terms is grounded in their relevance to the Blade Element Momentum theory (BEM) solution \cite{bauersfeld2021neurobem, kaufmann23champion}, reflecting their significant impact on the system dynamics while maintaining computational efficiency.} \remove{The respective coefficients are identified using force and torque measurements that are directly derived onboard IMU motorspeed and IMU measurements as well as from ground truth data, captured with a VICON system.}\rebuttal{The ground truth force and torque values are derived from IMU measurements, as well as from a VICON\footnote{\url{https://www.vicon.com}} system. The respective coefficients $\vec{C}_f$, $\vec{C}_\tau$ are then identified using ordinary linear least-squares regression, and remain constant at runtime.}
The data is gathered from the same race track to ensure that it captures the unique aerodynamic effects that arise from the racing maneuvers.
\remove{
It’s important to note that the coefficients are identified offline and remain constant at runtime.
Additionally, the residual model requires an accurate estimate of the motor speeds, which are not provided in the original MPCC formulation from Problem\mbox{~\ref{eq:full_ocp}}. To address this, we modfy the initial problem to include a first-order motor model within the nominal dynamics as outlined in Eq. \mbox{\eqref{eq:quad_dynamics}}}

\subsection{TuRBO tuning}
\label{sec:turbo}

We consider the MPCC++ controller as a black-box function in the context of \remove{Bayesian Optimization}\rebuttal{BO}. For a given set of controller parameters $\vec{\phi}$, we run an episode, collect a trajectory $\tau(\vec{\phi})$, and compute the reward $R(\tau(\vec{\phi}))$. For the sake of clarity, we \remove{will omit the trajectory from the notation and} simply refer to the reward as $R(\vec{\phi})$. The controller tuning task can be framed as an optimization problem:
\begin{equation}\label{eq:BOOpt}
\vec{\phi}_{n+1} = \arg\max_{\vec{\phi} \in \Theta} R(\vec{\phi})
\end{equation}
Problem \ref{eq:BOOpt} presents two caveats: i) no analytical form of $R(\vec{\phi})$ exists, allowing only for pointwise evaluation; \rebuttal{and} ii) each evaluation corresponds to completing an entire episode within the simulator, which means the number of evaluations is capped by our interaction budget with the simulated environment. 
BO works in an iterative manner, relying on two key ingredients: i) a probabilistic surrogate model approximating the objective function; and ii) an acquisition function $\alpha(\vec{\phi})$ that determines new evaluation points, balancing exploration and exploitation. 
A common choice for the surrogate model are Gaussian Processes (GPs), which allow for closed-form inference of the posterior mean $\mu(\vec{\phi})$ and variance $\sigma^2(\vec{\phi})$. 
The next evaluation point $\vec{\phi}_{n+1}$ is then chosen by maximizing the acquisition function $\alpha(\vec{\phi})$. 
We utilize the Upper Confidence Bound (UCB) acquisition function, defined as $\alpha_{UCB}(\vec{\phi}) = \mu(\vec{\phi}) + \beta \sigma(\vec{\phi})$, where $\beta$ is the exploration parameter. 
This process is iterated until the evaluation budget is exhausted. 
Among the various BO methodologies, we opt for TuRBO \cite{turbo}, a global BO strategy that runs multiple independent local BO instances concurrently. 
Each local surrogate model enjoys the benefits of local BO such as diverse modeling of the objective function across different regions. 
Each local run explores within a \remove{\textit{Trust Region}}\rebuttal{Trust Region} (TR) - a polytope centered around the optimal solution of the local instance. 
The base side length of the TR, $L$, is adjusted based on the success rate of evaluations to guide exploration towards promising areas. 
In this work we use the TuRBO-1 implementation from the BOTorch library \cite{balandat2020botorch} and modify it for an eightfold setup, TuRBO-8.

In contrast to the baseline MPCC \cite{mpcc}, which requires $N_{\vec{\phi}}=2n_{gates}+4$ parameters to be tuned due to the complexity of the contour function, MPCC++ reduces the number of tunable parameters to $N_{\vec{\phi}}=8$. From Problem \ref{eq:full_ocp}, the parameters in question within the MPCC++ \rebuttal{formulation} include $Q_l$, $Q_c$, $Q_\omega$, $R_{v_\theta}$, $R_{\Delta f}$ and $\mu$, with $Q_c$, $Q_\omega$ further divided into horizontal and vertical components.

For our experiments, we define one episode of the tuning task as the completion of $M=3$ consecutive laps around the track. The reward attained at the end of the episode is calculated as the mean lap time, adjusted by a penalty factor for any solver failures:
\begin{equation}\label{eq:rew}
    R(\vec{\phi}) = -\frac{1}{M}\sum_{i=0}^Mt_i-\gamma r_{fail}
\end{equation}
where $t_i$ denotes the lap time of the i-th lap, while $r_{fail}$ represents the solver failure rate, i.e. the proportion of steps that fail relative to the total steps. We apply a scaling factor of $\gamma=100$ to the penalty term. Such failures arise mainly due to \remove{infeasibilities when the system is pushed to its constraint bounds.}\rebuttal{i) the complexity of the optimization problem at hand; and ii) the need to solve the optimization problem in real-time (one SQP iteration).} Incorporating this penalty term\rebuttal{, while not strictly necessary,} has proven effective in practice, as it smoothens the objective function in favor of parameters that minimize solver failures. This reward structure is consistently applied across all experiments.

%% file: sections/experiments.tex
\section{Experiments}
In this section, we compare our proposed method, MPCC++, against two baselines: MPCC and RL, and conduct a series of \remove{ablation studies}\rebuttal{variations in the formulation}. 
The methodology is consistent across all experiments, and utilizes the same quadrotor configuration. 
We choose the Split-S race track, featuring 7 gates, for all our experiments due to its prevalent use in previous works \cite{Song23Reaching, CPC, mpcc, replanning}. 
We test \remove{the performance of }our approach across three distinct simulation environments: i) a simple simulator which uses the nominal dynamics; ii) a high-fidelity simulator that calculates the propeller forces via Blade Element Momentum Theory (BEM) \cite{bauersfeld2021neurobem}; and iii) a data-driven simulator that predicts aerodynamic forces from real-world data. 
We then validate our method in the real world. All experiments were \remove{executed}\rebuttal{conducted} on the same hardware under uniform conditions.

\input{figures/tables/main_table}
\subsection{Simulation}

We train the policies across the three simulators previously described using TuRBO as the default tuning method. Each policy is allocated a total budget of 600 episodes for training, which translates to a maximum of 1800 interactions with the environment, given that each episode involves completing 3 laps.
For the environment setup, simulation and training, we use a combination of \emph{Flightmare} \cite{yunlong2020flightmare} and \emph{Agilicious} \cite{foehn2022agilicious} software stacks.

At test time, we select the best policy from the training phase \rebuttal{as the one with the highest reward,} and run 10 episodes (equivalent to 30 laps). The results reported in Table \ref{tab:sim_real_results} are based on the average lap time over the 10 episodes \rebuttal{evaluated at test time}. \remove{Additionally, we report the success rate (SR) by counting the number of episodes where the drone successfully finishes all 3 laps without any gate collisions \mbox{\cite{Song23Reaching}}.}
\rebuttal{Additionally, we introduce the Training Success Rate (TSR) to quantify the percentage of training episodes completed successfully without gate collisions (out of 600). We also \remove{use}\rebuttal{employ} the Success Rate (SR) metric, as introduced in \cite{Song23Reaching}, to quantify the percentage of successful episodes during testing (out of 10).}

\remove{After simulation, we select the best policy obtained in the residual simulator and deploy it in the real world. Note that while BEM and the residual simulator provide comparable accuracy levels, the latter is preferred due to its significantly lower computational demand - approximately one tenth that of the BEM simulator.}

\subsubsection{TuRBO vs WML}
\begin{figure}[ht]
\centering
\vspace{-7pt}
\includegraphics[width=\linewidth]{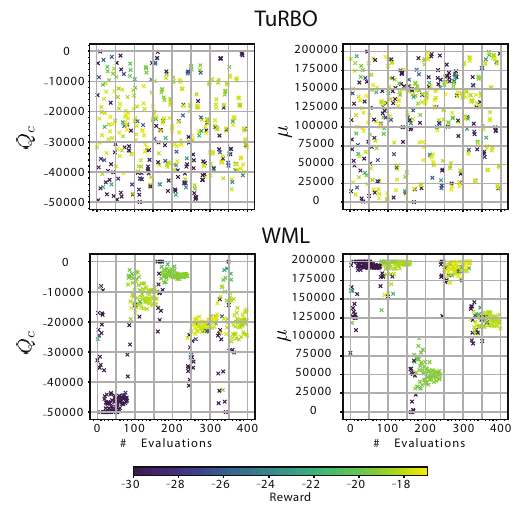}
\caption{\remove{Parameter exploration using WML vs TuRBO. The different data samples are colored based on the collected reward. The figure shows that the WML method is prone to get stuck in local minima, while the TuRBO method is able to explore a broader parameter space.}\rebuttal{Parameter exploration using TuRBO and WML. We plot two of the tuning parameters - $Q_c$ and $\mu$ - over the number of episodes. TuRBO's trust regions enhance exploration of the entire parameter space, while WML randomly restarts the exploration every 100 episodes.}}
\vspace{-7pt}
\label{fig:wml_vs_turbo}
\end{figure}

We first investigate the difference between \remove{WML and TuRBO}\rebuttal{TuRBO~\cite{turbo} and WML~\cite{wml}}, and discuss several advantages of TuRBO. We train the baseline MPCC \cite{mpcc} using both variants. Despite both methods achieving comparable maximum rewards, in this section we \remove{deive}\rebuttal{delve} into understanding how each of these explore the parameter space.\remove{ In} Figure~\ref{fig:wml_vs_turbo} \remove{we illustrate}\rebuttal{illustrates} two of the tuned parameters - the minimum contour weight $Q_c$ and the progress weight $\mu$ - \remove{across}\rebuttal{over} the number of episodes.

\remove{WML starts a new trial every 100 iterations, while TuRBO runs multiple trust regions in parallel, facilitating a broader exploration of the parameter space. This is primarily due to TuRBO's ability to  select the local instances in an informed manner, as opposed to WML's approach of random trial restarts.}\rebuttal{To encourage exploration, WML initiates a new trial every 100 iterations with random initialization, aiming to discover new parameter regions. In contrast, TuRBO employs a more strategic approach by selecting trust regions in a similar manner to how it selects BO points, using informed reasoning.} An additional advantage of TuRBO is that the trust regions are independent, allowing parallel execution without incurring extra computational cost. Consequently\rebuttal{,} we advocate for TuRBO as our preferred tuning algorithm\remove{,} for two reasons: i) comprehensive exploration of the entire parameter space; \rebuttal{and} ii) computational efficiency and scalability, which are crucial for exploring large parameter spaces.
\subsubsection{MPCC++ vs MPCC}
We train both \remove{MPCC and MPCC++}\rebuttal{MPCC++ and the baseline MPCC} using TuRBO. We make some practical adjustments to the MPCC++ formulation introduced in Section \ref{sec:tunnel_constraints}. Specifically, we relax the hard constraints of the tunnel and impose soft constraints instead.
\rebuttal{While our MPCC++ optimization problem provides numerical stability in settings that do not require real-time computations, employing real-time solvers does not inherently ensure numerically stable solutions. Even if a solution is found, real-world implementation presents challenges in ensuring the subsequent state respects the predefined constraints. This is particularly true when planning near boundaries, such as tunnel edges, where minor discrepancies due to model mismatches, disturbances, or state-estimation drift can lead to violations of these constraints in the subsequent states. Soft constraints are commonly employed to address such scenarios} and can be performed in a systematic way~\cite{wabersich2023predictive}. This has several benefits: i) improves the numerical stability of the solver; ii) handles infeasibilities which arise from model mismatch by allowing minor violations; \rebuttal{and} iii) maintains the low computational complexity of MPCC. This \remove{adaption does not impose any drawbacks}\rebuttal{adaptation does not have any practical implications} in terms of performance or safety. The soft constraint is implemented using a barrier function $p(h(x))$ to embed the constraint \rebuttal{$h(x)\geq0$} into the cost function: \remove{For nonlinear constraints defined as $h(x)\geq0$, we employ:}

\begin{equation} \label{eq:barrier}
    p(h(x)) = \log\left(1 + \exp\left(-\alpha h(x)\right)\right)
\end{equation}
\remove{setting}\rebuttal{where} the penalty slope \rebuttal{is set} to $\alpha=100$. Our results, as detailed in Table \ref{tab:sim_real_results}, demonstrate MPCC++'s superiority over MPCC in terms of lap times and success rates. 
Due to \remove{spatial constraints \mbox{\eqref{eq:tunnel_constraints}}}\rebuttal{the track constraint}, MPCC++ \remove{can prevent}\rebuttal{prevents gate} crashes consistently without compromising the lap time. 
Allowing the controller to independently navigate within the tunnel, instead of following a predefined path, gives it greater flexibility to identify the best trajectory based on its current state.
\remove{This strategy is akin to RL, enabling the controller to autonomously identify its optimal path, hence synthesizing the tasks of planning and control.}
The tunnel dimensions\remove{, determined by the nominal and inner gate values}, $W_n$ and $W_{gate}$, provide a mechanism to intuitively trade off \remove{safety against performance}\rebuttal{between performance and safety}. 
This balance was adjustable in test runs without necessitating retuning, indicating the policy's robustness across varying tunnel widths.

\begin{figure}[ht]
\centering
\includegraphics[width=\columnwidth]{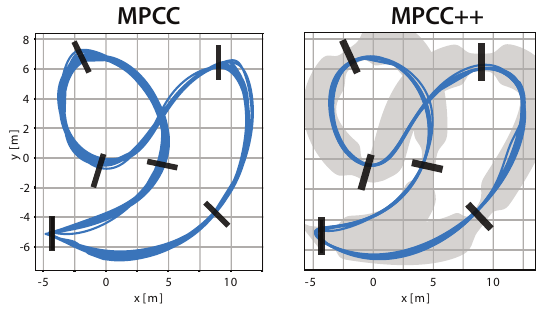}
\caption{Simulation experiments of MPCC with the proposed MPCC++, both tuned using TuRBO.}
\vspace{-7pt}
\label{fig:mpcc_vs_tmpc_sim}
\end{figure}

In Figure~\ref{fig:mpcc_vs_tmpc_sim}, we show the trajectory from 10 episodes (30 laps) for both MPCC and MPCC++, noting MPCC++'s paths are notably more consistent.
We attribute the inconsistency of MPCC to its representation of the contour weight as a multivariate Gaussian\remove{ introduces numerical asymmetry in the presence of mismatch}, since the controller forcefully tries to follow the reference path, and acts poorly in the presence of disturbances.

\remove{We introduce the Episode Success Rate (ESR) metric to quantify episodes completed without gate collisions.} As shown in Table \ref{tab:bo_racing}, MPCC achieves a \remove{ESR}\rebuttal{Training Success Rate (TSR)} of roughly 70\% across all simulations, indicating that 30\% of the episode evaluations are not completed successfully due to gate collisions, \rebuttal{while} MPCC++ \remove{approaches}\rebuttal{reaches} a \remove{ESR}\rebuttal{TSR} of nearly 100\%. Failed episodes are assigned with the lowest attainable reward and do not provide valuable insights for the BO's surrogate model updates. As such, we consider these as fruitless interactions with the environment.
\vspace{7pt}
\input{figures/tables/bo_table}

\subsubsection{MPCC++ with Model Augmentation}
Following the approach described in Section \ref{sec:dynamics_augmentation}, we augment the nominal dynamics of the MPCC $\vec{f(x,u)}$ with a residual component $\vec{g(x,u)}$.
\remove{
Since the residual term is a function of $v_x$, $v_y$, $v_z$, $\Omega^2$, requiring to adapt our MPCC++ scheme to model the motor speeds $\Omega_i$, rather than the direct thrusts $f_i$. To this end, we model the motor dynamics as a first-order system and define the states and inputs as described in Section \ref{sec:dynamics_augmentation}. In this case, increasing the number of states does not have a significant impact on the solver time in practice. 

It's also worth noting that the motor speeds are generally unavailable at runtime. We thus resort to estimating these online given the current state and applied command. These can be easily estimated from the motor model and low-level betaflight controller. Note also that including the residual terms in the MPC formulation potentially introduces more numerical instability, as the features are only plausible within a specific range. To prevent this, we impose an additional constraint on the motor speeds and velocity magnitude to prevent the features from estimating beyond the range in which they characterize the dynamics accurately.
}
\rebuttal{
Incorporating polynomial features to augment the dynamics requires knowledge of the motor speeds, which were not present in the original MPCC framework. To address this, we introduce several adaptations.

First, we include a first-order motor model into the dynamics to estimate the motor speeds in real-time:
\begin{equation}
    \dot{\vec{\Omega}} = \frac{1}{\tau_{mot}} (\vec{\Omega_{des}} - \vec{\Omega})
\end{equation}
where $\vec{\Omega_{des}}$ and $\vec{\Omega}$ denote the desired and the actual motor speeds, respectively.
Second, we require an accurate estimate of these speeds to initialize the controller at each timestep. Additionally, we reformulate the optimal control problem to use motor speeds as inputs instead of single rotor thrusts. These adaptations are essential for leveraging data-driven dynamic features within a real-time optimization framework, setting our approach apart from prior works like \cite{kaufmann23champion}, which used augmented dynamics for environment simulation in RL.

\remove{Moreover, including the residual terms in the MPCC formulation introduces potential numerical instability, as the features are valid only within a specific range. To mitigate this, we impose additional constraints on motor speeds and velocity magnitude to ensure the features do not extrapolate beyond their accurate characterization of the dynamics.}
}
\remove{We find that}\rebuttal{As shown in Table~\ref{tab:sim_real_results},} we are able to further reduce the lap time on all of the environments by approximately 0.1s. We attribute this improvement to a combination of the above: i) respecting the motor dynamics; and ii) the actual augmentation.
\subsubsection{MPCC++ with Domain Randomization}
We address the robustness of the control policy against different noise realizations. The goal is to find policies which perform well amid model discrepancies. Such mismatches arise from using e.g. different hardware, leading to slight changes in mass, inertia, or thrust. We adapt the existing training pipeline to account for such noise realizations. For each \rebuttal{BO} iteration \remove{of the BO loop}, we execute 10 different episodes each subjected to a distinct noise realization. We then collect the average reward over the 10 episodes and update the BO surrogate as usual. For fairness in comparison, we maintain the same \remove{number of interactions with the environment}\rebuttal{budget of episode evaluations}, which implies that the number of BO iterations is reduced by a factor 10 (from 600 to 60). In practice, reducing the number of BO iterations did not show any major limitations. We compare the obtained policies against the nominal MPCC++ and observe: i) slight increase in the lap time; ii)\remove{ in both cases} the consistency of the trajectory is preserved; \rebuttal{and} iii) 10 noise realizations are not sufficient for a truly robust policy. We conclude that the control formulation at hand\remove{ is inherently robust and} is able to adapt to both changes in the platform, as well as sudden changes in the dynamics. We owe this property to the controller's replanning capability within the constraint boundaries.

\subsubsection{MPCC++ vs RL}
\remove{
As shown in Table \ref{tab:sim_real_results}, our MPCC++ strategy attains lap times comparable to the leading RL policies \cite{yunlong2021racing}. We refer to \cite{Song23Reaching} for a detailed comparison between \remove{Optimal Control}\rebuttal{optimal control} vs RL. We discuss the difference between the trajectories \remove{of both in real-world} in Section \ref{sec:real_world}, and provide some general remarks herein. Despite both policies achieving similar performance, MPCC++ ensures safety by design. We also find that it's possible to fly with the same MPCC++ policy through a variety of tracks and achieve competitive lap times without retuning, indicating good generalization, while this is generally not the case for RL policies. It's worth noting that these advantages come at the price of more engineering efforts than an end-to-end learning framework. In particular, the need to account explicitely for delay compensation, solver infeasibilities, adapting the MPC outputs to low-level controller commands, alongside managing real-time computational solver limitations.
}
\rebuttal{
In this section, we compare our MPCC++ method against the RL policy from~\cite{Song23Reaching}. We refer to the cited work for a detailed comparison between optimal control and RL. As shown in Table~\ref{tab:sim_real_results}, both MPCC++ and RL achieve similar lap times. However, MPCC++ incorporates explicit safety considerations into its design, enabling it to optimize trajectories by strategically taking shortcuts through gate corners. In practice, MPCC++ employs a low contour weight, $Q_c$, which, due to the explicit safety constraints in its design, allows the solver to focus on planning the optimal trajectory without excessively penalizing the contour error. In contrast, RL penalizes the gate passing distances in its reward function, which encourages the drone to navigate through the gate center. A detailed comparison of the trajectories is provided in Section \ref{sec:real_world}.

While MPCC++ incorporates specific safety measures into its design, it requires considerable engineering effort compared to an end-to-end learning framework like RL. This includes managing real-time computational constraints of the solver, adapting MPC outputs to low-level controller commands, addressing solver infeasibilities, and explicitly compensating for delays. 
}

\begin{figure*}[ht]
\centering
\includegraphics[width=\textwidth]{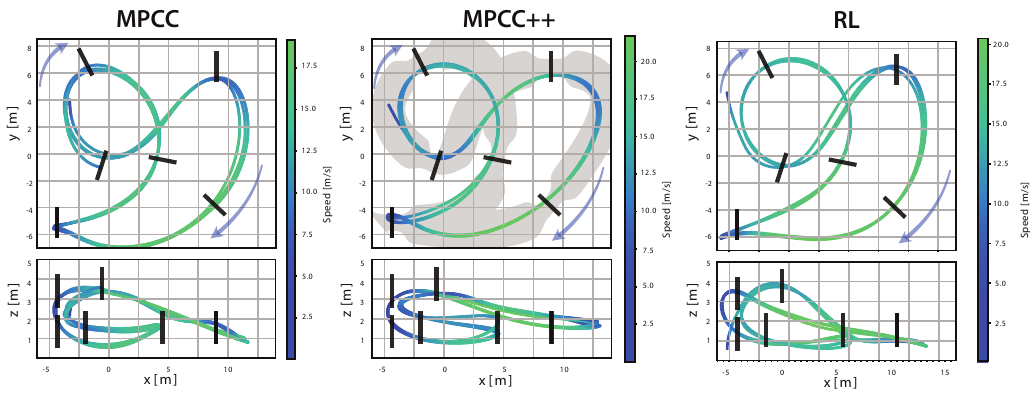}
\caption{\remove{Real-world flight comparison of the MPCC controller with the Tunnel MPC (TMPC) controller proposed in this work. Even when both controllers have been tuned with TuRBO, we show that the TMPC is able to achieve faster speeds, more consistency than its MPCC counterpart and 100\% success rate with no gate crashes.}\rebuttal{Real world flight trajectories on the Split-S track for the baseline MPCC~\cite{mpcc}, MPCC++ and the best-performing RL policy~\cite{Song23Reaching}. Both MPCC and MPCC++ were tuned using TuRBO.}}
\vspace{-7pt}
\label{fig:tmpc_vs_cmpc_real}
\end{figure*}
\vspace{10pt}
\subsection{Real World}
\label{sec:real_world}
We test our approach in the real world using a high-performance racing drone with a high thrust-to-weight ratio (TWR). We use the Agilicious platform \cite{foehn2022agilicious}\remove{ for the real-world deployment}, as \remove{detailed}\rebuttal{introduced} in \cite{Song23Reaching} under the designation \emph{4s drone}. The control framework was deployed on ACADOS\remove{\footnote{\url{https://github.com/acados/acados}}}~\cite{Verschueren2021} using SQP\_RTI for real-time computation, with the control loop \remove{runs}\rebuttal{running} at 100Hz. We use a horizon length of $N=20$ at a prediction rate of 25Hz, resulting in a prediction span of $T=0.8s$. 

Our method runs on an offboard desktop computer equipped with an Intel(R) Core(TM) i7-8565U CPU @ 1.80GHz. A Radix FC board \remove{that contains}\rebuttal{equipped with} the Betaflight\footnote{\url{https://www.betaflight.com}} firmware is used as \remove{a}\rebuttal{the} low-level controller\remove{. 
The low-level controller}\rebuttal{, which} takes as inputs \rebuttal{desired} body rates and collective thrusts. 
An RF bridge is employed to transmit commands to the drone. 
For state estimation, we use a VICON\remove{\footnote{\url{https://www.vicon.com}}} system with 36 cameras that provide the platform with millimeter accuracy measurements of position and orientation at a rate of 400 Hz.

\rebuttal{We select the best policy obtained in the residual simulator and deploy it in the real world.}
We compute the lap times and success rates as done for the simulation experiments, i.e. averaging over 10 \remove{trials (3 laps per trial)}\rebuttal{episodes}. In Figure \ref{fig:tmpc_vs_cmpc_real}, we show the comparison between\remove{ the real-world runs of} the baseline MPCC, MPCC++\remove{ (ours)} and the best\rebuttal{-performing} RL \rebuttal{policy}. Both MPCC and MPCC++ are tuned with TuRBO. MPCC++ consistently satisfies the tunnel bounds, while achieving higher speeds than\remove{ the} MPCC. Our policy did not crash a single time\remove{ during the 10 trials}, being the first approach to achieve a 100\% success rate in real-world \rebuttal{experiments}. The \remove{increase in robustness}\rebuttal{improvement in safety} comes without a compromise in performance, as our approach achieves similar lap times to the best\rebuttal{-performing} RL \remove{controller}\rebuttal{policy}. MPCC++ plans a trajectory \remove{very close to the edge of the gates}\rebuttal{that takes shortcuts through the gate corners}, while in RL the deviation from the gate center is penalized in the reward, hence encouraging the drone to fly closer to the gate centers. The Split-S maneuver at $x=-4.3m$ \rebuttal{and $y=-5.6m$}, stands out as a critical test of each approach's characteristics, as it's performed differently in all three cases. \rebuttal{This complex maneuver requires the drone to fly through a higher gate and then immediately descend through a second gate located directly below the first, with both gates sharing the same $x,y$ coordinates.} This is the most challenging maneuver of the track\remove{ and has a significant influence on}\rebuttal{, significantly influencing} the overall lap time. 

\begin{figure}[ht]
\centering
\vspace{-7pt}
\includegraphics[width=0.85\columnwidth]{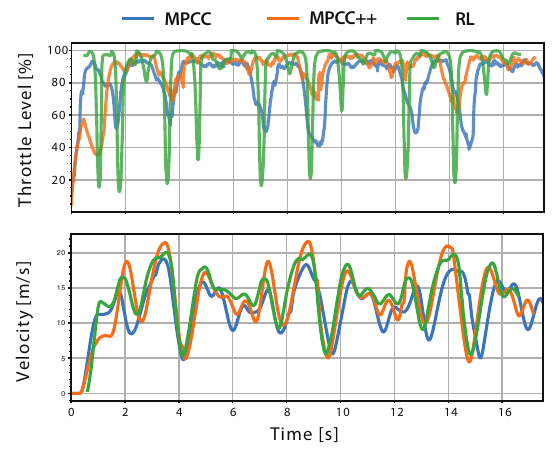}
\caption{Real world thrust and velocity profiles for MPCC, MPCC++ and RL. \remove{We note how the proposed approach is able to keep the platform extremely close to the limit of handling.}}
\vspace{-7pt}
\label{fig:throttle_real}
\end{figure}
Figure~\ref{fig:throttle_real} \remove{further explores}\rebuttal{illustrates} the differences in throttle level, representing normalized commanded collective thrust, and velocity magnitude among the approaches. While velocity profiles were similar, the commanded throttle values differ significantly\remove{ between the MPC approaches and RL}. This is because \remove{in MPC}\rebuttal{for MPCC and MPCC++}, the solver output needs to be translated to equivalent desired body rates and collective thrust values \remove{, which are then}\rebuttal{to be} commanded to the low-level controller. This mapping is non trivial, and needs to account for delay effects. We believe that it has a significant influence on the overall performance and is a major drawback of MPC\rebuttal{-based} approaches\remove{. On the other hand}, \rebuttal{while} RL directly outputs a suitable command\remove{ for the low-level controller}. 
\section{Discussion}
This paper introduces MPCC++, a safe MPCC controller for drone racing\remove{that prevents collisions against the gates}. 
\rebuttal{We first introduce safety constraints, in the form of a track constraint and a terminal set, which systematically prevent gate collisions and ensure recursive feasibility.}
We \rebuttal{then} augment the nominal dynamics of the controller with a residual term that\remove{ is obtained from real-world data and} captures unmodeled behaviors such as aerodynamic effects\remove{ and propeller forces}.
\remove{We}\rebuttal{Finally, we} tune the controller parameters using TuRBO, a state-of-the-art \remove{Bayesian optimization}\rebuttal{BO} algorithm that efficiently compromises between local and global exploration. 
Our approach addresses several of the limitations encountered in the baseline MPCC\remove{ \cite{mpcc}}: i) \remove{robustness against gate collisions}\rebuttal{no systematic prevention against gate collisions}; ii) sub-optimality from following a predefined optimal trajectory; \rebuttal{and} iii) tuning effort incurred from\remove{ tuning} a high dimensional parameter space.
We benchmark \remove{our approach}\rebuttal{MPCC++} against the baseline MPCC and the best-performing RL \remove{controller}\rebuttal{policy} in both simulation and real-world experiments and show that we are on par in terms of lap times while attaining safety against \rebuttal{gate} collisions. Our approach is the first to achieve 100\% success rate in real-world experiments.

Nonetheless, MPCC++ also presents several limitations. 
First, \remove{there is a need for}\rebuttal{it requires} a centerline that passes through the gate centers and is used as the basis to construct the \remove{safety set}\rebuttal{track constraint}.  
\remove{While there is no computational effort incurred from it, as it can be almost any smooth spline, it does require some manual effort to be determined.
Additionally, unlike RL architectures, our approach restricts the platform to operate within a predefined safety set. 
This limitation hinders its ability to freely explore and develop new behaviors necessary for executing complex maneuvers, such as loops or ladders, which are commonly seen in drone racing environments.
}
\rebuttal{
While constructing the centerline incurs no computational effort, this approach does introduce certain limitations.
Specifically, our approach requires mapping the drone's current position to a point on the centerline, which is then linked to a progress value. 
Additionally, constructing the centerline as a spline requires defining the task as a sequence of waypoints in a known order.
In contrast, RL architectures often use reward functions (sparse, binary, etc) to specify tasks, allowing for a simpler and more flexible task specification.
}
\remove{It remains for future work to understand how the choice of the centerline influences the solution and find a method that generalizes to all track configurations.}

Second, our approach takes significantly longer to train than RL, as the environment cannot be parallelized on the GPU. This is because the solver calls cannot be batched into tensor operations, as done with \remove{MLP}\rebuttal{learning-}based policies.

\remove{Third, the performance of our controller is limited at run time as it requires an optimization problem to be solved at each step, whereas RL methods require a simple forward pass of the neural network.}
\rebuttal{Third, while our approach assumes that gate positions are known with high accuracy, future work should explore incorporating gate position uncertainty into our method.}

\rebuttal{Finally, our approach requires solving an optimization problem online at each step, which may limit its applicability on onboard computers with limited processing capabilities. In contrast, RL methods only require a simple forward pass of the neural network at runtime.}

\remove{Thus, we}\rebuttal{We} conclude that the main advantage of \remove{our approach}\rebuttal{MPCC++} is that it provides an intuitive lever to trade off between performance and \remove{robustness}\rebuttal{safety}. By shrinking the width of the tunnel, the quadrotor flies safely through the track. We can then gradually adjust the width to increase performance. Adjustments in the tunnel width can be made directly without \remove{further fine-tuning}\rebuttal{the need of further re-tuning}.

\section{Acknowledgments}
This work was supported by the European Union’s Horizon Europe Research and Innovation Programme under grant agreement No. 101120732 (AUTOASSESS) and the European Research Council (ERC) under grant agreement No. 864042 (AGILEFLIGHT). The authors especially
thank Amon Lahr for his helpful acados tips, and Jiaxu Xing and Ismail Geles for their help with the experiments.

%% file: figures/tables/main_table.tex
\begin{table*}[ht]
    \centering
    \vspace*{-7pt}
    \setlength{\tabcolsep}{2pt}
    \begin{tabular}{c|c|c|cc|cc|cc|cc}
    \toprule
    \multirow{3}*{\textbf{Category}} &  \multirow{3}*{\textbf{Methods}}&  \multirow{3}*{\textbf{Tuning}} & \multicolumn{8}{c}{\textbf{Environments}}\\
    &&&\multicolumn{2}{c|}{\textit{Simple}}& \multicolumn{2}{c|}{\textit{BEM}}& \multicolumn{2}{c|}{\textit{Residual}} & \multicolumn{2}{c}{Real World}\\
    &&&Lap Time [s]&SR[\%]& Lap Time [s]&SR[\%]& Lap Time [s]&SR[\%]& Lap Time [s]&SR[\%]\\
    \midrule
    \multirow{2}*{MPCC \cite{mpcc}}& \multirow{2}*{Nominal}
    & WML \cite{wml} & 5.38 $\pm$ 0.1 & 100 & 5.51 $\pm$ 0.06 & 100 & 5.51 $\pm$ 0.13 & 83.3 & - & -\\
    && TuRBO & 5.65 $\pm$ 1.07 & 89.7 & 5.37 $\pm$ 0.06 & 100 & 5.62 $\pm$ 0.23 & 96.7  & 5.67 $\pm$ 1.06 & 59.3\\
   \midrule
   \multirow{3}*{MPCC++ (ours)}
   & Nominal & TuRBO & 5.16 $\pm$ 0.02 & 100 & 5.30 $\pm$ 0.02 & 100 & 5.37 $\pm$ 0.09 & 100 & 5.41 $\pm$ 0.14 & 100\\
   & w/ augment. & TuRBO & 5.09 $\pm$ 0.10 & 100 & 5.15 $\pm$ 0.03 & 100 & 5.19 $\pm$ 0.03 & 100 & 5.38 $\pm$ 0.26 & 100\\
   & w/ random. & TuRBO & 5.20 $\pm$ 0.13 & 100 & 5.37 $\pm$ 0.08 & 100 & 5.26 $\pm$ 0.27 & 100 & - & - \\
   \midrule
   \multirow{1}*{RL \cite{Song23Reaching}} & - & - & 5.14 $\pm$ 0.09 & 100 & - & - & 5.26 $\pm$ 0.32 & 100 & 5.35 $\pm$ 0.15 & 85.0 \\
    \bottomrule
\end{tabular}
    \caption{\remove{Results for MPCC, MPCC++ and RL. There are four columns depending on which environment the experiment was conducted in. Three of these columns are for different simulation environments, and the last column for real world results. We can see how for our MPCC++, the success rate in all simulations and in the real world is 100\%.} \rebuttal{Results for MPCC, MPCC++, and RL across three simulation environments - Simple, BEM, and Residual - as well as for real-world experiments. MPCC++ achieves a 100\% success rate across all simulation and real-world environments.} \label{tab:sim_real_results}}
    \vspace*{-10pt}
\end{table*}

%% file: figures/tables/bo_table.tex
\begin{table}[ht]
    \centering
    \setlength{\tabcolsep}{2pt}
    \begin{tabular}{c|c|c|c|c|c}
    \toprule
    \multirow{2}*{\textbf{Category}} &  \multirow{2}*{\textbf{Methods}}&  \multirow{2}*{\textbf{Tuning}} & \multicolumn{3}{c}{\textbf{Environments}}\\
    &&&\multicolumn{1}{c|}{\textit{Simple}}& \multicolumn{1}{c|}{\textit{BEM}}& \multicolumn{1}{c}{\textit{Residual}}\\
    \midrule
    \multirow{2}*{MPCC \cite{mpcc}}& \multirow{2}*{Nominal}
    & WML \cite{wml} & 62.9 & 75.7 & 64.4\\
    && TuRBO & 69.2 & 70.3 & 53.0\\
   \midrule
   \multirow{3}*{MPCC++ (ours)}
   & Nominal & TuRBO & 99.5 & 100 & 99.8\\
   & w/ augment. & TuRBO & 100 & 100 & 100\\
   & w/ random. & TuRBO & 91.3 & 93.9 & 89.0\\
    \bottomrule
\end{tabular}
    \caption{BO \remove{Exploration} \rebuttal{Training} Success Rate \rebuttal{(TSR)}: percentage of episodes during training \remove{which successfully complete all laps}\rebuttal{completed successfully} without gate collisions.
    \label{tab:bo_racing}}
\end{table}